\documentclass[10pt,conference,letter,final]{IEEEtran}
\usepackage{etex}

\usepackage{cleveref}
\crefname{figure}{fig.}{figs.}
\Crefname{figure}{Fig.}{Figs.}
\crefname{equation}{}{}
\crefname{section}{Section}{Sections} \crefname{subsection}{Section}{Sections} \Crefname{subsection}{Section}{Sections} 
\usepackage{amsmath}
\usepackage{gensymb}
\usepackage{psfrag}
\usepackage{graphicx}
\usepackage{pgf}
\usepackage{url}
\usepackage{cite}
\usepackage{listings}
\usepackage{units}
\usepackage[autolanguage]{numprint}
\usepackage{booktabs}
\usepackage[xindy,toc]{glossaries}
\usepackage{bytefield}
\usepackage{xspace}
\usepackage{microtype}
\usepackage{tikz}
\usepackage{tikz-timing}

\usepackage{textcomp}
\usepackage[binary-units,detect-all]{siunitx}

\usepackage{changes}
\usepackage[colorinlistoftodos,obeyDraft]{todonotes}

\newcommand {\tdi}[1]{}

\robustify{\gls}
\robustify{\glspl}

\nprounddigits{2}

\makeglossaries

\newacronym{adc}{ADC}{Analog to Digital Converter}
\newacronym{dac}{DAC}{Digital to Analog Converter}
\newacronym{ppu}{PPU}{Plasticity Processing Unit}
\newacronym{isa}{ISA}{Instruction Set Architecture}
\newacronym{fifo}{FIFO}{First In First Out}
\newacronym{fpga}{FPGA}{Field Programmable Gate Array}
\newacronym{fwhm}{FWHM}{full width at half maximum}
\newacronym{serdes}{SerDes}{Serializer/Deserializer}
\newacronym{simd}{SIMD}{Single Instruction Multiple Data}
\newacronym{sram}{SRAM}{Static Random Access Memory}
\newacronym{stdp}{STDP}{Spike-timing dependent plasticity}
\newacronym{io}{IO}{Input/Output}
\newacronym{dls}{HICANN-DLS}{HICANN-DLS}
\newacronym{lsb}{LSB}{least significant bit}
\newacronym{inl}{INL}{integral nonlinearity}
\newacronym{msb}{MSB}{most significant bit}
\newacronym{gpu}{GPU}{Graphics Processing Unit}
\newacronym{asic}{ASIC}{Application Specific Integrated Circuit}
\newacronym{hicann}{HICANN}{High Input Count Analog Neural Network}
\newacronym{hicann-figure}{HICANN}{HICANN}
\newacronym{cmos}{CMOS}{Complementary Metal-Oxide-Semiconductor}
\newacronym{nmda}{NMDA}{N-Methyl-D-Aspartat}
\newacronym{nmda-nx}{NMDA}{NMDA}
\newacronym{psp}{PSP}{post-synaptic potential}
\newacronym{ota}{OTA}{operational transconductance amplifier}
\newacronym{adex}{AdEx}{Adaptive-Exponential Integrate-and-Fire}
\newacronym{hbp}{HBP}{Human Brain Project}
\newacronym{hpc}{HPC}{High Performance Computing}
\newacronym{pp}{PP}{plateau potential}

\makeatletter
\renewcommand*{\@seccntformat}[1]{   \csname the#1\endcsname.\quad
}
\makeatother

\newcommand{\greatTitle}{An Accelerated Analog Neuromorphic Hardware System Emulating NMDA- and Calcium-Based Non-Linear Dendrites}
\title{\greatTitle}

\author{
	\IEEEauthorblockN{Johannes~Schemmel\IEEEmembership{Member,~IEEE},
	Laura Kriener, Paul M\"uller and Karlheinz~Meier}
	\IEEEauthorblockA{Heidelberg University, Heidelberg, Germany
	\\\{schemmel,lkriener,pmueller,meierk\}@kip.uni-heidelberg.de}
}
\begin{document}
\maketitle

\begin{IEEEkeywords}
    neuromorphic hardware, multi-compartment neuron, neuron circuit, non-linear dendrites, NMDA
\end{IEEEkeywords}

\maketitle

\begin{abstract}
This paper presents an extension of the BrainScaleS accelerated analog neuromorphic hardware model. 
The scalable neuromorphic architecture is extended by the support for multi-compartment models and non-linear dendrites.
These features are part of a \SI{65}{\nano\meter} prototype \gls{asic}.
It allows to emulate different spike types observed in cortical pyramidal neurons: NMDA plateau potentials, calcium and sodium spikes. 
By replicating some of the structures of these cells, they can be configured to perform coincidence detection within a single neuron.
Built-in plasticity mechanisms can modify not only the synaptic weights, but also the dendritic synaptic composition to efficiently train large multi-compartment neurons.
Transistor-level simulations demonstrate the functionality of the analog implementation and illustrate analogies to biological measurements.
\end{abstract}

\section{Introduction\label{sec:introduction}}
The biological nervous system is a main source of inspiration in the quest for future computing. A prominent example is deep learning, a computing scheme based on multi-layer Perceptron models \cite{lecun2015deep}, which is currently in the focus of academia as well as industry \cite{jones2014}. The neuron models used in these machine learning applications are heavily simplified compared to the biological example. The fact that such a simple copy of the basic architecture of the nervous system is already capable of impressive results encourages many scientists that even more powerful computing devices might be built by looking more closely at nature's principles. The hope is that even without a full understanding of the operation of the brain, studying its architecture leads to new inspirations for the development of novel computing systems\cite{hawkins2016}.

Testing such concepts is mostly done with numerical simulations, often using standardized software packages \cite{nestinitiative08homepage, neuronhomepage2}. These tools allow a step up in complexity from the Perceptron model by using spike-based neuron models. Spike-based models cover a wide range of complexities, ranging from the basic Integrate-and-Fire models up to Hodgkin-Huxley-like models incorporating a multitude of different ion-channel kinetics \cite{markram2006blue}. 

It has been shown by several research groups that prominent concepts used with Perceptron models, for example sampling based approaches \cite{petrovici2016stochastic, neftci2015event} or deep-learning using back-propagation \cite{rumelhart86backprop}, can be transferred successfully to spiking models\cite{bothe2000}. Comparing typical benchmark problems, many spike-based implementations reach similar scores as their rate-based examples \cite{hunsberger2015}, but they execute orders of magnitude slower on the commonly used \gls{hpc} platforms\cite{helias2012supercomputers}.

 It gets even worse if the training methods are constrained to biologically plausible mechanisms, which requires that at short timescales all information exchange is done by spikes only. This forbids for example the implementation of back-propagation as it is currently used in the standard deep-learning software packages \cite{tensorflow2015}. Luckily, first ideas how to circumvent these problems have been reported lately \cite{nokland2016}, where the network learns not only its objective but also the correct mapping of the error information from the output backwards to the upstream synapses.

Recent findings in biology \cite{antic2010nmda} have inspired novel models \cite{hawkins2016,schiess2016} which include the three-dimensional structure of the neuron. The dendritic tree is no longer treated as a compartmentalized cable-equation \cite{brown1975}, but as a complex non-linear structure which allows multi-layer information processing and coincidence detection within a single neuron \cite{larkum2009}.

Including all these details into numeric simulations strongly enhances the performance problems already surfacing with spiking neuron models. The work in this paper presents an alternative approach to the numerical modeling of multi-compartmental, non-linear dendrites, using physical-model-based neuromorphic hardware\cite{indiveri2011}. It builds upon the BrainScaleS accelerated analog neuromorphic system \cite{schemmel_iscas2012} in conjunction with the built-in hybrid plasticity  extension developed for the BrainScaleS~2 system\cite{friedmannschemmel2016}. It expands these concepts by incorporating novel circuits to mimic non-linear dendritic behavior, including an emulation of \gls{nmda} and calcium channels\cite{schiller2000nmda}. 

In \cite{qiao2015reconfigurable} a neuromorphic chip is presented that also contains \gls{nmda} emulation circuitry and has adapted many of the features of the BrainScaleS system, but implementing them in the real-time domain using sub-threshold point neurons, in contrast to the accelerated emulation used in the circuits presented in this publication. Some authors have reported neuromorphic circuits incorporating aspects of \gls{nmda}-R behavior to achieve a certain functionality, for example \cite{rachmuth2011biophysically} and \cite{you2016neuromorphic}. In contrast to those, our approach is not targeted at a single functional model, but aims to be a universal experimental platform. The presented implementation is also inherently scalable in the framework of the BrainScaleS system. Neuromorphic hardware concepts based on non-linear dendrites have also been previously reported \cite{banerjee2015current}\cite{hussain2016morphological}, demonstrating the viability of the concept for pattern classification.

This paper will present an extension of the BrainScaleS \gls{adex} neuron model \cite{touboul08adex} that allows the replication of coincidence detection between basal and apical dendritic segments similar to those observed in experiments \cite{larkum2013}. The circuits are fully plastic and can be tuned during the experiment according to plasticity rules executed by the local plasticity processing unit \cite{friedmann13phd}, while still performing the network emulation at an acceleration factor of $10^3$. A first prototype chip has been sent to manufacturing at the time of this writing.

The remaining sections of the paper are organized as follows: Section~\ref{sec:brainscales2} gives an introduction to the BrainScaleS~2 accelerated analog neural network hardware implementation, Section~\ref{sec:mcneurons} describes the theory as well as the circuit implementation of the multi-compartment extensions. Section~\ref{sec:results} shows simulation results demonstrating the presented circuits' capabilities. Section~\ref{sec:plasticity} discusses the built-in plasticity and possible learning algorithms. The paper closes with Section~\ref{sec:outlook}, presenting our conclusions and outlook.

\section{Accelerated Analog Neuromorphic Hardware\label{sec:brainscales2}}

\begin{figure}
    \center{\includegraphics[width=0.98\linewidth,page=1,viewport=0 0 20.6cm 13.8cm, clip]{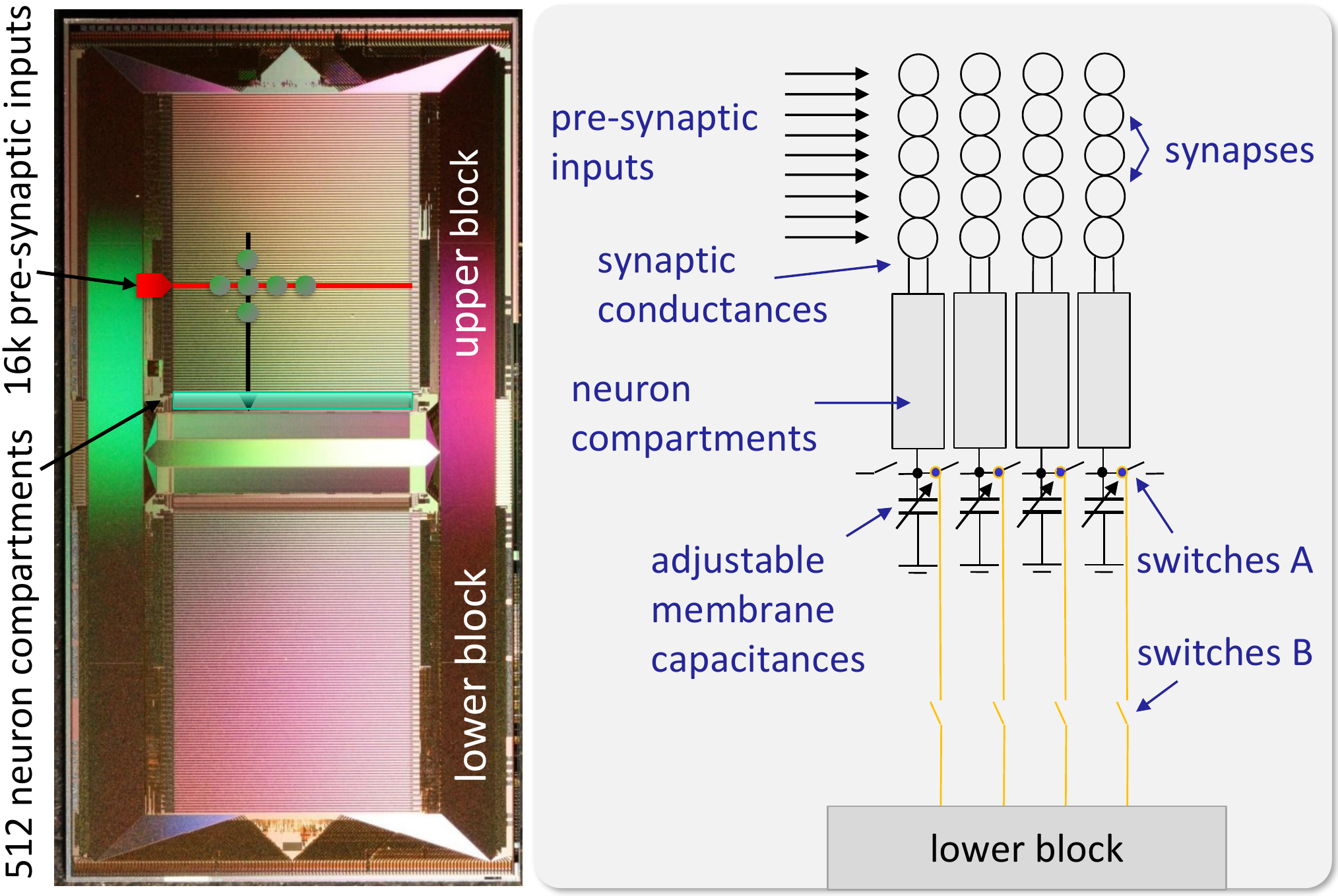}  
    \caption{Basic structure of the HICANN chip. Left: Micro-photograph of a first generation \gls{hicann-figure} chip. The synapses are organized in two blocks with the neurons in the center. Each block has 128 independent pre-synaptic input circuits, each capable of relaying 64 different pre-synaptic neurons, adding up to a total of 16k different pre-synaptic neurons. A single input is symbolically depicted by the red arrow. Right: conceptual drawing of the neuron compartment circuits and their associated synapses. The individual compartments are connected to each other by switches to assemble the neurons from the compartments. There are switches between neighboring compartments (A) and between adjacent compartments in the upper and lower block (B).
    \label{fig:hicannblock} }}
\end{figure}

The presented multi-compartment circuits are part of a larger research project which aims to develop the second generation BrainScaleS neuromorphic hardware as part of the European \gls{hbp}\cite{Markram2011introducing}. The basic neuromorphic building block of every BrainScaleS system is the \gls{hicann} chip. It contains the neuron and synapse circuits as well as a digital communication network. While the first generation is implemented in \SI{180}{\nano \meter} standard \gls{cmos} technology, the second generation uses a smaller \SI{65}{\nano \meter} feature size, which enables, among others, the inclusion of a \gls{ppu} to implement hybrid plasticity \cite{friedmannschemmel2016}.

Fig.~\ref{fig:hicannblock} shows the basic structure of the BrainScaleS neuromorphic \gls{asic}. The micro-photograph is the current version of the BrainScaleS chip and serves only as an illustration, since the basic structure of the second generation BrainScaleS \gls{asic}, which is the version referred to in this paper, will be very similar. The synapses are arranged in a two-dimensional array. All synapses in a column share their output, while two adjacent rows share the same group of pre-synaptic input signals. There are 512~rows all-together, each group of two is connected to 64 pre-synaptic neurons by the means of the digital communication network. The inputs to the upper and lower block are independent from each other. Each block can receive events from a maximum of 8192 different pre-synaptic neurons. 

To be able to create neurons of different sizes, each column of synapses together with the neuron compartment circuits in the center of the chip has an adjustable membrane capacitance which can be connected to the neighboring compartment circuit by an electronic switch\footnote{All switches are built from standard \gls{cmos} transmission gates.}. A second set of switches allows to connect adjacent neuron compartments in the upper and lower block, doubling the number of available pre-synaptic inputs to the neuron. The maximum number of pre-synaptic neurons that can project to a single neuron is 16k. The pre-synaptic neurons can be located on the same or on remote chips, either on the same or different silicon wafers \cite{schemmel_iscas2010}.

\begin{figure}
    \center{\includegraphics[width=1\linewidth,page=2,viewport=0 0 21.6cm 15.9cm, clip]{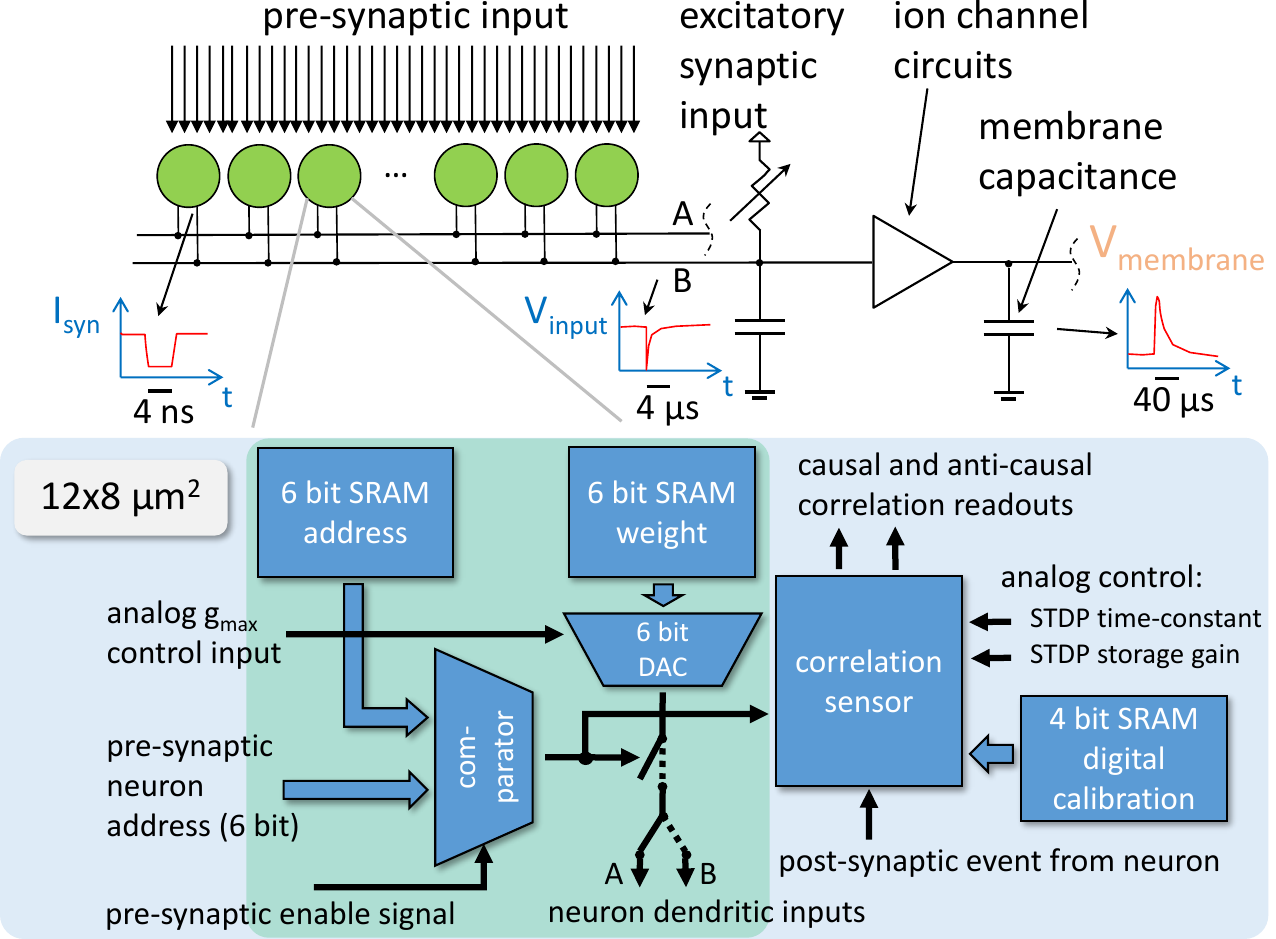}  
    \caption{Conceptual drawing of the synaptic input with its associated column of synapses (rotated by 90 degrees to fit the layout of the figure), the enlargement shows a block diagram of the synapse circuit. The part of the synapse implementing the pre-post connection is shaded in green.
    \label{fig:synapticinput} }}
\end{figure}

Fig.~\ref{fig:synapticinput} illustrates the synaptic input of the neuron with its associated synapses as well as the temporal relationship of the related signals. It shows that a column of synapses (rotated in the figure to fit the page format) is connected to the two \emph{dendritic input lines} of the neuron compartment, labeled \emph{A} and \emph{B}. The enlargement in the lower half of Fig.~\ref{fig:synapticinput} depicts the different functional blocks within each synapse. At its core is a \SI{6}{\bit} memory storing the current weight of the synapse. A weight of zero means there is effectively no connection between the pre- and post-synaptic neurons, but the correlation between pre- and post-synaptic events is still being monitored\footnote{This will be covered in detail in Section~\ref{sec:plasticity}.}.

\begin{figure}
    \center{\includegraphics[width=0.95\linewidth,page=3,viewport=0 0 23.4cm 15.3cm, clip]{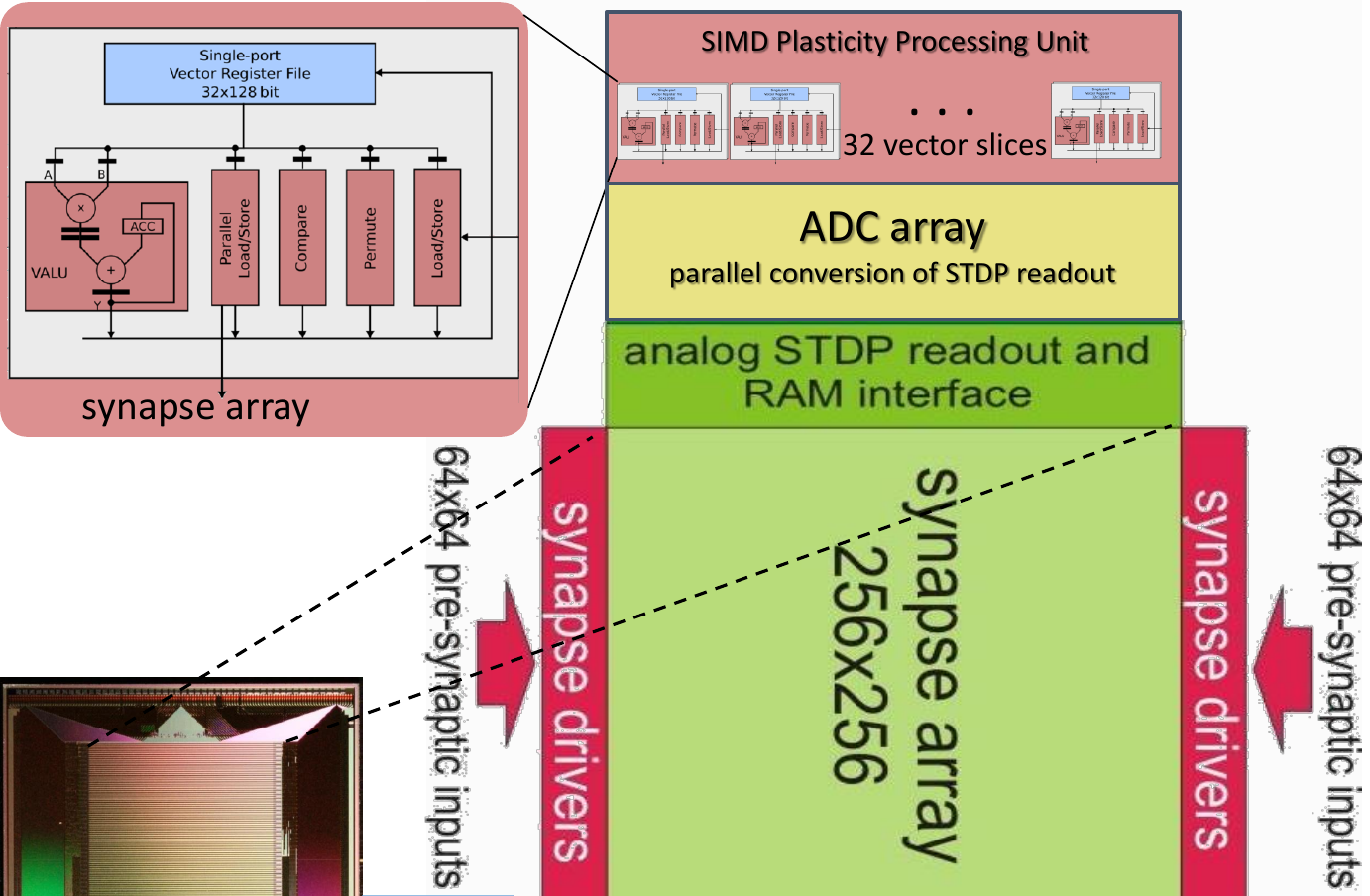}  
    \caption{One \gls{ppu} instance each is located a the outer edge of the upper and lower synapse block.
    \label{fig:ppu} }}
\end{figure}

The neuron compartment circuit emulates the different ion channels. The voltage on the membrane capacitance reflects the momentary membrane voltage of the compartment\cite{millner10}. The conductances and capacitances are scaled such that all time constants are a factor of 1000 shorter compared to biology. Hence the addition \emph{accelerated} in the designation of the BrainScaleS model. 

A time-multiplexed scheme is used to allow the high number of inputs per row. The communication network delivers pre-synaptic events with a maximum rate of \SI{125}{\mega \hertz}. Each row of synapses receives pre-synaptic events from up to 64 different pre-synaptic neurons through said network. The synapses receive the events via one shared input bus per row, transmitting a \SI{6}{\bit} pre-synaptic address to identify the pre-synaptic neuron. Each synapse stores a \SI{6}{\bit} pre-synaptic address that is compared to the address presented on the input bus each time a pre-synaptic event is transmitted. In case of an address match, the synapse uses a pulse of a precise duration of \SI{4}{\nano \second} to sink current from the dendritic input of the neuron compartment circuit it is connected to. The amplitude of the current pulse is proportional to the weight stored in a separate \SI{6}{\bit} memory within the synapse\cite{friedmannschemmel2016}. Depending on a row-wise configuration setting, the synapses use one of two available dendritic inputs, named \emph{A} and \emph{B} in Fig.~\ref{fig:synapticinput}. This allows each neuron compartment to accommodate two different synaptic time constants and reversal potentials. The capacitance of the dendritic input line acts as an integrator of all synaptic current pulses. An adjustable resistor recharges the dendritic line capacitance continuously, thereby setting the synaptic time constant. Due to the acceleration factor the synaptic time constant is typically about \SI{2}{\micro \second}, approximately three orders of magnitude slower than the synaptic current pulses. 

By storing not only their weight, but also part of their pre-synaptic neuron address, the content of the synapse memories defines the local network topology. To program these memories, a custom \gls{simd} micro-processor (\gls{ppu}) is connected to the synapse array at the opposite edge of the compartment circuits. Fig.~\ref{fig:ppu} illustrates this arrangement. It is described in detail in \cite{friedmannschemmel2016}.

\section{Multi-Compartment Neurons\label{sec:mcneurons}}
\subsection{Conceptual Background\label{sec:mcintro}}

\begin{figure}
    \center{\includegraphics[width=0.65\linewidth,page=4,viewport=0 0 9.2cm 16.2cm, clip]{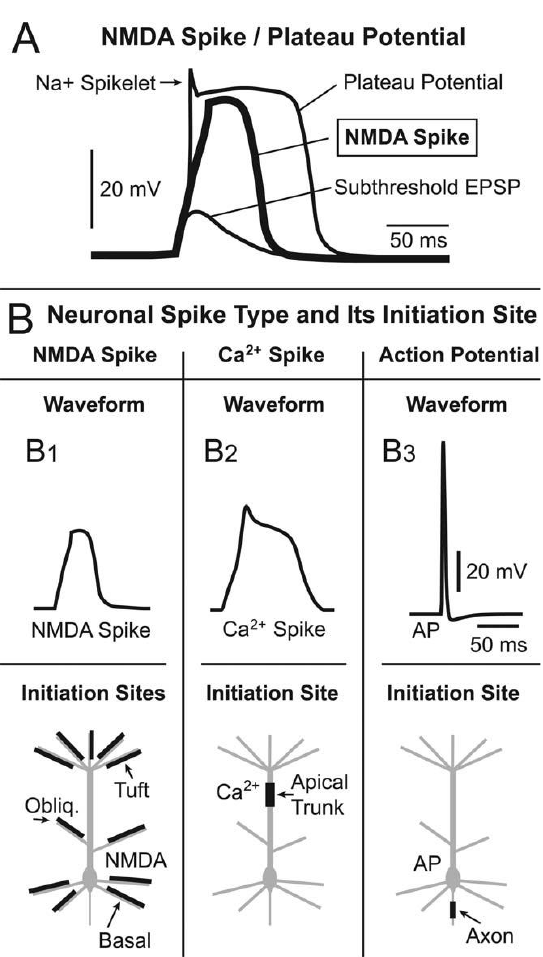}  
    \caption{Different kinds of spikes in a cortical pyramidal neuron. Figure taken from \cite{antic2010}.
    \label{fig:anticspikes} }}
\end{figure}

The accelerated analog neuron model presented in Section~\ref{sec:brainscales2} can be used to emulate point-neuron-based network models with a biologically realistic fan-in of more than 10k pre-synaptic neurons.
To use a physical model with such a high number of inputs one has to consider the linear character of the synaptic input in the point neuron model. Identical synapses generate the same \glspl{psp}, thus having the same potential contribution to the firing of the soma. A contribution which decreases with the total number of synapses, since in the BrainScaleS physical model increasing the size of the neuron automatically decreases the \gls{psp} of a single synapse. This is caused by the growth of the membrane capacitance due to the larger number of neural compartment circuits connected together.

Therefore, the more synapses a neuron has, the more pre-synaptic action potentials must arrive in synchrony to reliably relay an input pattern. On the other hand it is desirable to use sparse coding in neural networks\cite{palm2013neural,kanerva2009hyperdimensional}. The energy consumption in the BrainScaleS physical model is directly linked to the sparseness of the neural code used.

Arguments against the simple linear addition of all synaptic inputs can also be found in biology. It has been observed that many neo-cortical cells are subject to a dense background firing from thousands of synapses as well as that microscopic sources of true noise are present at each synapse\cite{faisal2008noise}. 

Recent findings have shed some light on different non-linear mechanisms within the dendritic tree of the neuron\cite{london2005dendritic}, most notably the capability of the dendritic membrane to generate different spike types in distinct parts of the dendritic tree of a neuron. These properties provide a possible solution to the problem outlined above. A hypothesis reviewed in \cite{antic2010}: ``Clustering of synaptic inputs in space (and time) improves the chances for reaching the dendritic threshold for firing a regenerative (amplified) response and provides the opportunity for faster and more frequent cooperation among synaptic contacts involved in the same computational task.'' Therein the authors furthermore elaborate: ``Instead of thousands of synaptic inputs, the pyramidal cell requires only a \emph{correct} set of~$<$~50 active synaptic contacts to trigger a regenerative dendritic response (e.g.~NMDA/plateau potential)''. Fig.~\ref{fig:anticspikes}, taken from \cite{antic2010}, summarizes the three kinds of spikes one can observe within a cortical pyramidal neuron.

\gls{nmda}-receptors are typically located in the thin, distal parts of the tuft, oblique and basal dendrites. They are responsible for \gls{nmda} spikes ($B1$). Since they are usually co-located with sodium channels, the resulting waveforms resemble the \gls{nmda} \gls{pp} shown in Fig.~\ref{fig:anticspikes}A. Triggered by a sufficiently localized synaptic input of approximately 10 to 50 pre-synaptic action potentials\cite{losonczy2006integrative}, it strongly increases the membrane conductance for a period ranging from several tens to hundreds of milliseconds. Due to the fact that \gls{nmda} channels are glutamate receptors with voltage dependent magnesium blocks, the \gls{nmda} \gls{pp} is a strongly non-linear function of the pre-synaptic input. If the dendritic membrane stays below its threshold, only a sub-threshold \gls{psp} is observed.

The presented in-silico emulation has been guided by these observations.
Because the design of our physical modeling system already incorporates the interconnection of compartments to implement a scalable number of synapses per neuron it was a natural step to include an extension that incorporates dendritic structures with active components. In the subsequent sections we focus on pyramidal neurons becuase of their supposed role in cortical information processing, as discussed for example in \cite{larkum2013}.

 \subsection{Implementation\label{sec:mccircuits}}
The implementation of the multi-compartment concepts introduced in Section~\ref{sec:mcintro} into the BrainScaleS~2 neuron requires two additional features for the existing neuron circuit \cite{aamir16dlsneuron}: an emulation of the effect of the \gls{nmda} and calcium ion channels as well as controllable inter-compartmental conductances.
\begin{figure}
    \center{      \includegraphics[width=1.05\linewidth,page=9,viewport=0 0 20.9cm 7cm, clip]{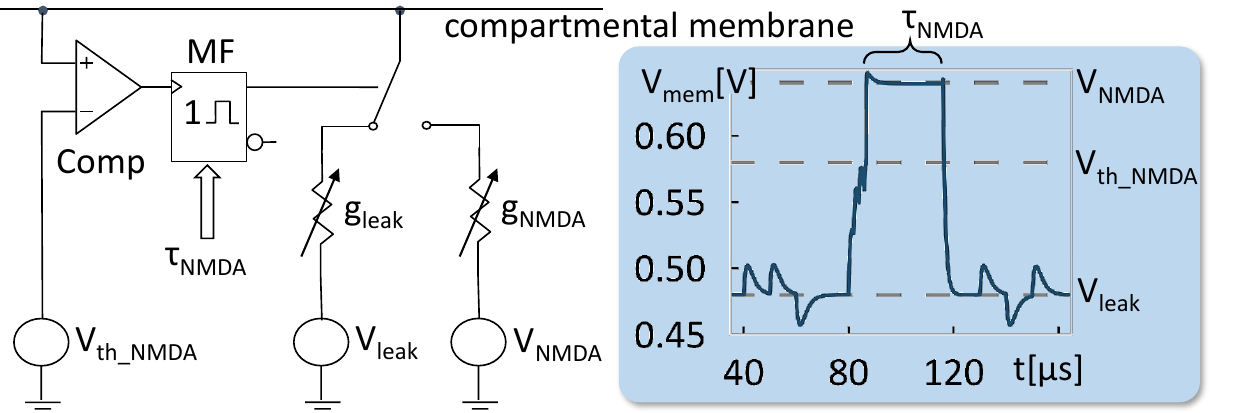}
      \caption{Left: Operation principle of the compartmental ion channel circuit (exemplarily configured as \gls{nmda} circuit). Right: Transistor-level simulation of an \gls{nmda} \gls{pp}.
    \label{fig:nmda} }}
\end{figure}
The previous neuron implementations of our research group, beginning with the Spikey neuron \cite{indiveri2011}, and including our previous multi-compartment chip \cite{millner12esann}, are only capable of emulating a sodium-like spike. This is done by continuously comparing the membrane voltage against an adjustable threshold voltage. If the threshold voltage is crossed, a spike is generated and the membrane is connected to the reset potential by a very high conductance. This condition is held for an adjustable amount of time to generate the refractory period of the neuron. After the refractory time has passed, the connection to the reset is released and the membrane is controlled by the interplay of synaptic input and leakage potentials again.

Fig.~\ref{fig:nmda} illustrates the operational principle of the ion channel circuit. It is based on a unified emulation circuit for the three different neuronal spike types listed in Fig.~\ref{fig:anticspikes}. The ion channel circuit uses two adjustable settings for its reversal potential as well as its conductance. The active setting is controlled by a voltage comparator (Comp), which continuously compares the membrane voltage against an adjustable threshold. If the threshold voltage is crossed, the ion channel circuit switches to the alternate setting. The output signal of the comparator passes through a mono-flop (MF) which ensures that the ion channel circuit switches to its alternate setting for a defined period of time. In the presented implementation this time interval is controlled by a digital counter, allowing a wide dynamic range from sub-milliseconds to several hundreds of milliseconds in biological time. The ion-channel itself is built from an \gls{ota} circuit, emulating the channel conductance. Electronic switches connect one of the two electrical parameter sets to the \gls{ota}. The parameters are part of the analog parameter storage memory associated with the neuron compartment circuits\cite{hock13analogmemory}. This memory holds 24 analog parameters for each individual neuron compartment. \added{In addition to parameter tuning, the values stored in the analog memory are also used to compensate process and fixed-pattern variations.}

In Fig.~\ref{fig:nmda} the circuit is configured for the emulation of \gls{nmda} channels: the threshold is set to the gating voltage of the \gls{nmda} receptor ($V_\textrm{th\_NMDA}$), the ion channel emulation is switched from the leakage setting ($g_\textrm{leak}$ and $V_\textrm{leak}$) to the setting for an \gls{nmda} \gls{pp} ($g_\textrm{NMDA}$ and $V_\textrm{NMDA}$). The right half of the figure shows a transistor-level simulation of this circuit, demonstrating the effect of the voltage-gated \gls{nmda} channel on the membrane voltage: as soon as multiple \glspl{psp} pile up and reach the \gls{nmda} threshold, the conductance mode is switched and the \gls{nmda} conductance pulls the membrane quickly up to the \gls{nmda} reversal potential $V_\textrm{NMDA}$ where it stays until $\tau_\textrm{NMDA}$ has passed and the ion channel emulation switches back to the leakage parameters, pulling the membrane back to the compartment's leakage potential $V_\textrm{leak}$.

By changing the values of the threshold as well as the two parameter sets for the ion channel and the time constant of the mono-flop other kinds of spikes can be emulated as well. For example, in the sodium case the threshold equates to the firing threshold and the time constant to the refractory time. Instead of using the \gls{nmda} parameters the ion channel circuit is set to maximum conductance and the reversal potential to the reset voltage. This will be shown in more detail in Section~\ref{sec:results}. 

In the current revision of the BrainScaleS~2 chip, each neuron compartment circuit contains one instance of the functional unit described above. Therefore each compartment can now be configured to generate either \gls{nmda}, calcium or sodium spikes. All three spikes also generate digital signals that can be routed as events to other parts of the system, which are typically but not exclusively the pre-synaptic inputs of unrelated neurons. Thus, they also take part in the coincidence detection mechanisms used for plasticity (see Section~\ref{sec:plasticity}). Since all parameters are freely adjustable within the available ranges, settings which do not resemble biological examples can be realized as well. If no voltage-gated ion channels are required, the circuit can be disabled.

The presented models are still simplistic and do not take into account some of the known features of their biological examples. For example, the glutamate concentration at the distal dendrite modulates the length of the \gls{nmda} \gls{pp}\cite{antic2010nmda}. Also, the channel emulation is only voltage gated while the real \gls{nmda}-R molecule is a glutamate receptor with a voltage dependent magnesium block.

\begin{figure}
    \center{\includegraphics[width=0.85\linewidth,page=5,viewport=0 0 16.1cm 11.9cm, clip]{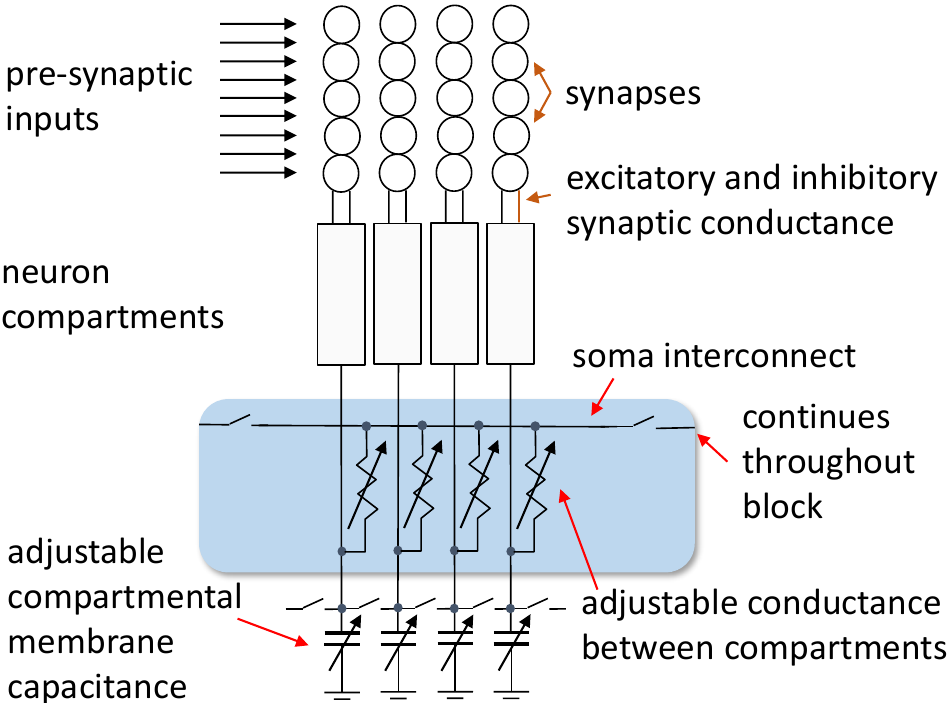}  
    \caption{Multi-compartment extension to the BrainScaleS neuron circuits. The additional parts are shaded.
    \label{fig:mcext} }}
\end{figure}

The second extension is an additional interconnect to create larger neurons from a set of neuron compartments. The shaded part in Fig.~\ref{fig:mcext} shows the necessary components: an adjustable conductance per neuron compartment and some switches. The new shared line represents the somatic membrane. Each neuron compartment can be connected to it via an adjustable conductance which represents the conductance between the distal dendrite and the soma. Usually, not all neuron compartments within a neuron block (see Fig.~\ref{fig:hicannblock}) are supposed to be part of the same neuron. Therefore, there are switches built into the somatic line at regular distances (every four compartments in Fig.~\ref{fig:mcext}) which allow its separation into different neurons. The somatic membrane by itself does not contain any active circuits, neither does it have an associated membrane capacitance. It acquires this functionality by a connection to a neuron compartment which is configured to have an infinite conductance between its compartmental membrane and the somatic membrane. To achieve the effect of this infinite conductance, or zero resistance, a bypass switch exists in parallel to each adjustable conductance.

\begin{figure}
  \centering
  \includegraphics[width=1\linewidth,page=6,viewport=0 0 16.4cm 19cm, clip]{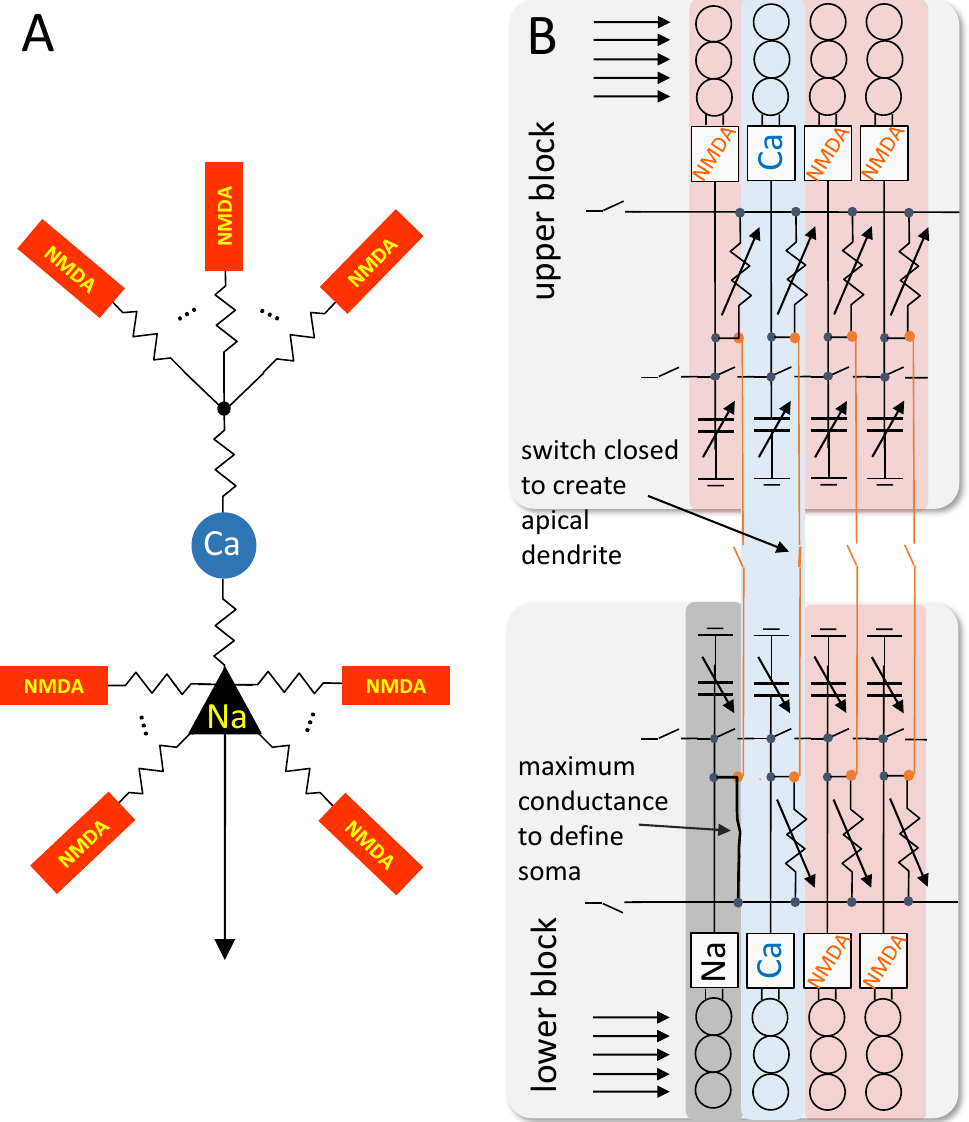}
  \caption{
    A: Illustration of the equivalent pyramidal neuron model in accordance to \cite{larkum2009}. The different compartments are discerned by their spike mechanism: \gls{nmda}, calicum (Ca) or sodium spikes (Na).
    B: Example configuration to implement the multi-compartment structure shown in A. The configuration of each compartment is depicted by its coloring as well as by the name of the spike type they produce (NMDA: red, Ca: blue, Na: black).
       \label{fig:mcpneuronandconfig}
     }
\end{figure}
Including said extensions into the basic neuron block structure illustrated in Fig.~\ref{fig:hicannblock}, the neuron can now be configured to emulate non-linear, multi-compartment neurons. By including calcium spikes it is possible to emulate more complex neurons, like for example layer 5 pyramidal neurons \cite{larkum2009}. Fig.~\ref{fig:mcpneuronandconfig}A illustrates this concept:
the basic structure of a pyramidal neuron is replicated using the presented circuits. The individual compartments are connected by the adjustable inter-compartment conductances introduced in Fig.~\ref{fig:mcext}, depicted by resistor symbols. The neuron model consists of a set of distal tuft and basal dendrites containing \gls{nmda} receptors, giving them the ability to create \gls{nmda} \gls{pp}s if the \gls{nmda} threshold is reached. All basal distal dendrites are connected to the soma, which is configured to generate sodium spikes to emulate the axon hillhock. The distal tuft dendrites converge at a separate junction, emulating the apical dendrite. The apical dendrite connects to the soma via a compartment configured for calcium spike generation. This allows the electronic neuron model to detect coincidences between its basal and distal input, similar to the measurements of layer 5 pyramidal neurons reported in\cite{larkum2013}. 
Section~\ref{sec:results} provides simulation results illustrating this mechanism.

Fig.~\ref{fig:mcpneuronandconfig}B depicts the same configuration of the neuron compartment circuits, but shown as they are arranged in the physical layout of the chip (see Fig.~\ref{fig:hicannblock}). Both neuron blocks, the upper and the lower, are used. The somatic line in the upper block emulates the apical dendrite, the one in the lower block the soma. The lower left compartment is configured to generate sodium spikes. Therefore, the bypass switch for the resistor connecting it to the somatic line is closed. Its membrane capacitance becomes the somatic capacitance of the neuron. If the voltage on this capacitance crosses the sodium threshold programmed into the compartment, the neuron will fire a spike and the soma capacitance will be pulled down to the reset potential. The \gls{nmda} compartments (marked red in Fig.~\ref{fig:mcpneuronandconfig}B) in the lower block emulate the distal basal dendrites, the ones in the upper block the distal tuft dendrites. 

Two compartments, one in the upper and one in the lower block, are used to connect the apical dendrite with the soma (shown in blue). This is accomplished by closing the vertical switch between them. This directly connects both their membrane capacitances and both calcium spike mechanisms share the same membrane voltage, which is isolated from the soma as well as the apical dendrite by an adjustable conductance each. The calcium spike generation can use either one or both of the spike mechanisms in the two compartments, which provides additional possibilities for better approximating the correct calcium waveform by combining multiple time constants and conductances. The possible neuron models are not limited to the pyramidal neuron example. For instance, it is also possible to create a branch in the apical dendrite, or to have several distinct dendrites.

\section{Results\label{sec:results}}

\begin{figure*}
  \centering{
    \includegraphics{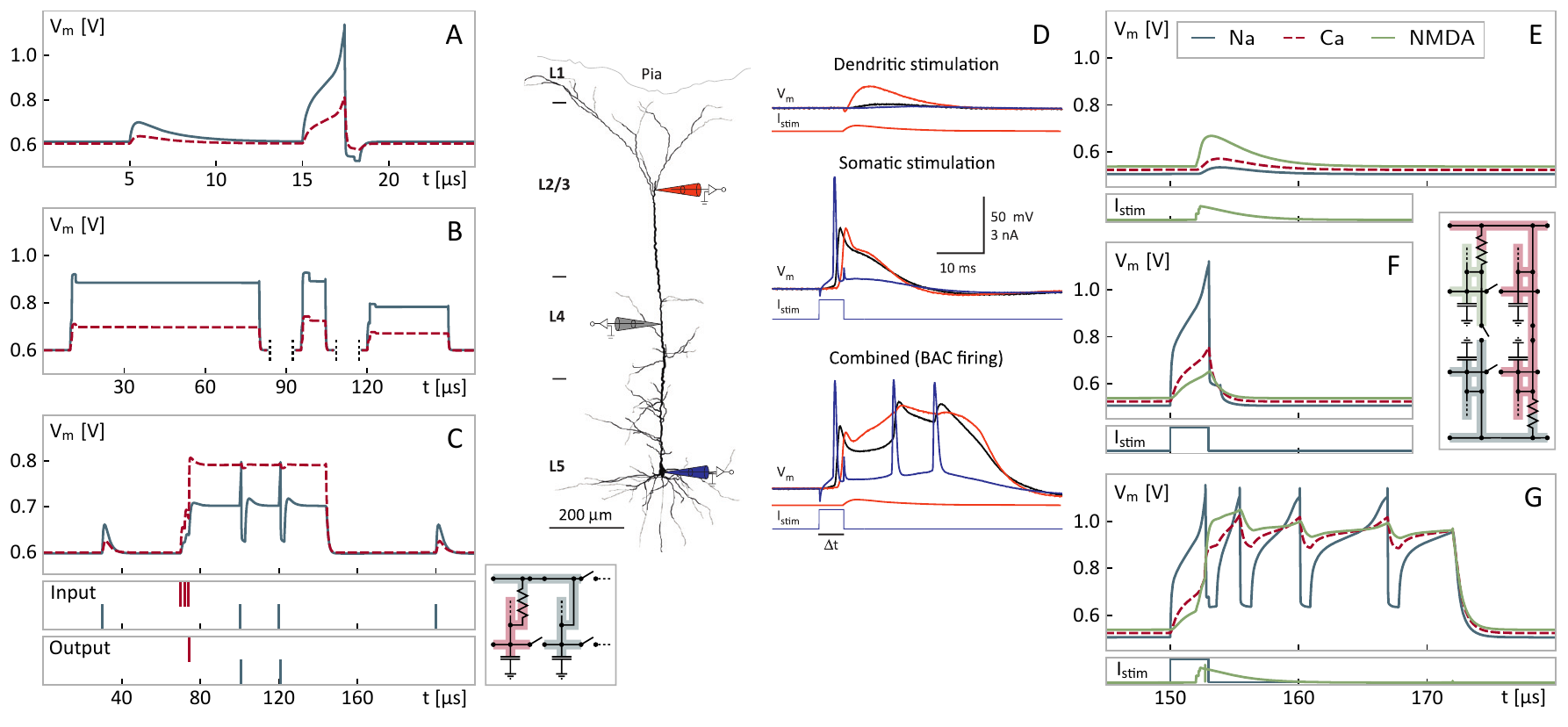}
  }
  \caption{
    {\bf Simulation of multi-compartment functionality:}
        A: A single compartment (solid line) is stimulated by synaptic
    input and a current pulse.
        The current pulse is sufficient to trigger the exponential current
    in the compartment.
        Consequently, the firing threshold is reached and the compartment
    is pulled to the reset voltage which is configured to be below
    \SI{0.6}{V}.
        The neighboring compartment is connected using an adjustable
    conductance (Fig.~\ref{fig:mcext}) and its membrane voltage
    (dashed line) passively follows the active compartment.
        B: Demonstration of the configurability of the reset circuit
    (Fig.~\ref{fig:nmda}).
        The reset duration and voltage can be modified to implement
    an up-state of precise length (solid lines).
        The neighboring compartments are pulled up via the passive
    conductance (dashed lines).
        Three distinct simulations for refractory periods of
    \SIlist{70;9;30}{\micro\second} are shown.
        C: Detection of coincidence over long timescales.
        Top: Synaptic input into the \emph{soma} compartment (solid line)
    only induces spikes when the neighboring compartment (dashed line) has
    entered a high state.
        Center: input spikes to \emph{dendritic} compartment (red, top) and soma
    (blue, bottom).
        Bottom: spike emission for
    dendritic (top) and soma (bottom)
    compartments.
        Bottom right: configuration of compartment interconnection for
    this simulation.
        {\bf Biologically inspired use case:}
    D: Figure taken from \cite{larkum2013}. The response of a layer~5
    pyramidal neuron to separate and correlated dendritic and somatic
    stimulus is shown. The traces show the membrane voltage at the Na-spike initiation zone (blue),
    the Ca-spike initiation zone (black) and the dendrites (red).
        E--G: Simulation of circuit configuration that is inspired by the
    reference in D.
        Only the stimuli differ in the simulations in E--G; the circuit
    configuration is identical.
        The amplitude of the current stimulus is \SI{1.5}{\micro\ampere}.
        E: Spike input into the NMDA compartment (green, bottom) is
    chosen as not to cause firing in any compartment.
        An attenuated version of the post-synaptic potential
    is seen in the Ca and Na compartments.
            F: Current stimulus into the Na compartment which is
    adjusted to initiate a single spike.
        The pictogram shows the switch and resistor configuration
    to which the simulations in E--G will correspond in the final chip.
        G: Both inputs combined suffice to cause firing in the Ca
    and NMDA compartments.
        This, in turn, induces a burst in the Na compartment.
        Note that the time scale is milliseconds for the biological
    reference (D) and microseconds for the circuit simulation (E--G) due
    to the accelerated nature of the neuromorphic device
    (Section~\ref{sec:brainscales2}).
        The circuit and stimulus parameters in all simulations were chosen
    such as to reproduce the desired spiking behavior.
        There is no procedural parameter mapping from a given reference
    for circuit or stimulus parameters.
        The relative timing of the stimulus in E--G is chosen such
    that the current input into the soma compartment arrives first,
    as in D.
        The time scales are not translated quantitatively.  }
\label{fig:simulation_results}
\end{figure*}

This section provides results from circuit simulations of the neuron and
its multi-compartment extensions.
The simulation setup includes four neuron compartments with the
corresponding multi-compartment circuits and eight synapses for each
compartment.
Only the part of the system that is essential to the new multi-compartment
functionality is simulated at transistor level to reduce computation
time. The simulated circuits match those of the prototype \gls{asic}.
 The Spectre simulator\footnote{Cadence Design Systems, Inc., San Jose, CA, USA}
is used with device characterization data provided by TSMC\footnote{Taiwan Semiconductor Manufacturing Company, Ltd., Hsinchu, Taiwan}
for the simulations.

A behavioral model is implemented for the mono-flop which triggers the
start of the refractory period and the alternate conductance mode
(Fig.~\ref{fig:nmda}) when it receives a signal from the spike
comparator.
In the chip, the digital configuration is stored in local
\gls{sram} while the analog parameter memory provides currents and
voltages to the respective circuits.
In simulation, the SRAM is implemented at transistor level and each
cell is initialized to the required value for the corresponding
neuron setup.
The analog parameter memory is simulated as a behavioral model which
consists of the output stage for current and an effective ideal
capacitance and resistor for voltage parameters \cite{hock13analogmemory}.

The chip design provides the possibility to stimulate one neuron by 
external current in addition to spiking input that reaches the
compartments via the synapse circuits.
In the simulation, current and voltage signals are provided to
implement these inputs to the neurons and synapses as the ideal
version of the input that is seen by the circuits during operation.
In particular, the pre-synaptic enable signal and neuron address
(Fig.~\ref{fig:synapticinput}) are provided to each synapse to
initiate synaptic events while the current stimulus is injected into
a shared input line.
One neuron at a time is configured to receive input from this line.

Voltage readout for an arbitrary compartment is possible via a
dedicated read-out path in the chip design.
This path is not included in the simulation to reduce simulation time
and the signals that are shown are recorded directly from the
corresponding capacitors in the circuit.

Fig.~\ref{fig:simulation_results}A shows the basic functionality of
a multi-compartment configuration.
One compartment receives two inputs of different strength.
The exponential term of the \gls{adex} implementation is enabled for
the compartment which receives the input.
This term generates a current onto the membrane that increases as an
exponential function of the membrane potential itself.
It acts as a soft threshold \cite{touboul08adex} in addition to the
explicit firing threshold in each compartment (Fig.~\ref{fig:nmda}).
The second input is sufficiently strong to cause the membrane voltage
of the stimulated compartment to exceed the threshold of the
exponential term in that compartment and induce a spike and reset.
The reset is configured to have a short refractory period and a low
reset voltage, which corresponds to typical point-neuron models with
Na spikes, e.g.\ the \gls{adex} model \cite{brette_05}.
The neighboring compartment is passively pulled up via the
inter-compartment conductance.
The upswing of the membrane voltage during an action potential is not
captured in the implemented circuit as is usual for low-dimensional
spiking neuron models (cf.\ \cite{gerstner2014dynamics}), which
is particularly beneficial for a hardware implementation as it allows
for a better utilization of the available voltage range for the neural
sub-threshold dynamics.

The refractory time, reset voltage and threshold can be configured
individually for each compartment
(Fig.~\ref{fig:simulation_results}B), which is central to the
implementation of active dendrites (Section~\ref{sec:mcneurons}).
Here, the reset potential is set above the threshold and the
reset duration is set to three values between \SI{9}{\micro\second}
and \SI{70}{\micro\second}.
Since the reset conductance is configured to be greater than the
leak conductance, this setting effectively serves as an additional
positive input current to the neighboring compartment,
which is being pulled up passively.

A demonstration of directed coincidence detection is shown in
Fig.~\ref{fig:simulation_results}C.
Two compartments, one with a Na-like (short reset) and one with
an NMDA-like (plateau potential) configuration, are connected
by a conductance and each compartment is stimulated by distinct
synaptic input.
The circuit parameters are adjusted in such a way that the Na
compartment emits spikes for single synaptic inputs during the high
state of the NMDA compartment, but does not when the
NMDA compartment is in its inactive state.

The features described above (Fig.~\ref{fig:simulation_results}A--C) are
used to implement a functional behavior which is similar to that of
layer 5 pyramidal neurons (Fig.~\ref{fig:simulation_results}D) described in
\cite{larkum2013}.
Therein its author hypothesizes that pyramidal neurons in the cortex act as coincidence detectors
for their basal and apical inputs.
This proposed mechanism employs the active nature of Ca and NMDA
spikes (Fig.~\ref{fig:anticspikes}) to allow a non-linear interaction
of synaptic input to opposite poles of the neuron.
Fig.~\ref{fig:simulation_results}D shows how a dendritic stimulus
leads to a marginal effect at the soma, while a somatic stimulus leads
to a single action potential.
Both inputs combined trigger a burst.
This functionality is emulated using the active components in the
presented circuits (Fig.~\ref{fig:simulation_results}E).
Synaptic
input to the NMDA compartment induces a \gls{psp} which
propagates to the other compartments and is attenuated along the way.
Current stimulus into the Na compartment
(Fig.~\ref{fig:simulation_results}F) is set to cause a single spike.
When both inputs are applied simultaneously the voltage in both
dendritic compartments crosses the respective threshold, pulling up
the membrane voltage in the Na compartment and causing a burst
(Fig.~\ref{fig:simulation_results}G).

This demonstrates how the presented physical implementation is
configured in analogy to a biological use case, emulating the
dendritic structure by a series of connected compartments and using
the introduced extensions of the BrainScaleS neuron model (Section~\ref{sec:mcneurons})
to emulate the active nature of the Ca and NMDA spikes
in the biological reference.
The simulation shows that using this structural analogy one can
parameterize the circuit to achieve a functional analogy, in this case
the implementation of a non-linear coincidence detection for inputs
into different locations of a single neuron.

\section{Plasticity\label{sec:plasticity}}
In Section~\ref{sec:mcneurons} the basic concept of the BrainScaleS accelerated analog neuromorphic network chips has been presented, omitting one important aspect: plasticity.
Similar to other neuromorphic devices, e.g.\ \cite{kasabov2013dynamic,pfeil2013neuromorphic}, correlation measurement between pre- and post-synaptic events is used to implement learning.
In Fig.~\ref{fig:synapticinput} and Fig.~\ref{fig:ppu} two key structures implementing plasticity\footnote{We are restricting this chapter to the multi-compartment related aspects of long-term plasticity. The BrainScaleS chips also implement short-term synaptic plasticity \cite{schemmel_iscas07}.} are shown: the correlation sensor within each synapse and the \gls{ppu} at the edge of each synapse array.
The correlation sensor measures the exponentially weighted temporal difference of each pre-post and post-pre spike pair\footnote{The correlation sensor implements a nearest-neighbors scheme.} and stores it locally in each synapse.
The \gls{ppu} can read back these causal and anti-causal correlation data as well as the current weights and addresses of the synapses. It executes a software-defined algorithm to determine new weights and possibly new addresses. It can also update all parameters of a configured neuron circuit like Fig.~\ref{fig:mcpneuronandconfig}, e.g.\ modify \gls{nmda} plateau durations, calcium threshold voltages or reset conductances. A detailed description of how these circuits interact to implement a flexible hybrid plasticity concept can be found in \cite{friedmannschemmel2016}.

Measuring correlation between pre- and post-synaptic events is frequently used to implement local learning in neuromorphic hardware \cite{kasabov2013dynamic}\cite{pfeil2013neuromorphic} based on the strengthening of causal connections, i.e. synapses that were active in the time frame before the firing of the post-synaptic event.
In large neurons with linear dendrites, this becomes increasingly difficult because of the diminishing effect a single synapses has on the firing of the post-synaptic event. The non-linear model using \gls{nmda} \gls{pp}s may provide better signatures for plasticity. Only about 10 to 50 synapses located on a common distal dendrite are needed to evoke a plateau potential\cite{losonczy2006integrative}. Therefore, replacing the \gls{nmda} event as the post-synaptic event for plasticity provides a clear learning signal even with thousands of synapses connected to the neuron. The presented hardware model allows to use all spike types as post-synaptic signals for plasticity, including \gls{nmda} for the synaptic columns configures as distal dendrites.

Grouping synapses onto distal dendrites with non-linear, active mechanisms solves the problem of the single synaptic event drowning in the overall synaptic input of the cell, but creates a new one: which synapses should be grouped on a dendrite? The presented neuromorphic hardware realizes an efficient platform for testing algorithms using the combination of local correlation measurement and the possibility to quickly change the pre-synaptic input of a dendrite. This implements a hardware analogy to the kind of structural plasticity created by the growth of axons and dendrites and the formation of axonal boutons and dendritic spines\cite{butz2013simple}\cite{knott2006spine}. 
\begin{figure} 
    \center{\includegraphics[width=1.02\linewidth,page=8,viewport=0 0 20cm 9.4cm, clip]{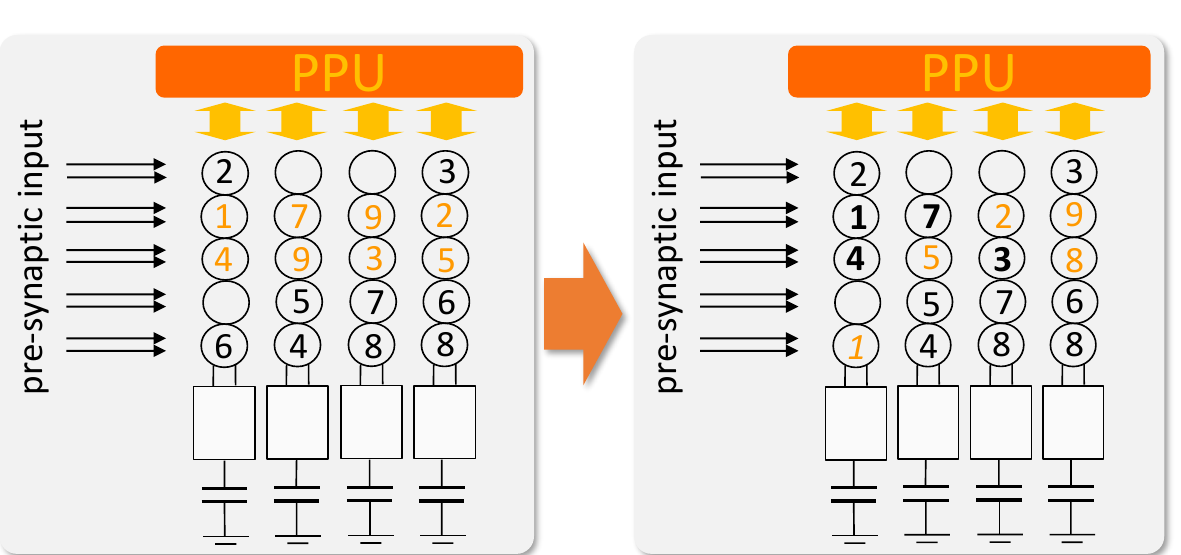}  
    \caption{Illustration of structural plasticity. Part of the pre-synaptic address memories in the synapses are changed by the \gls{ppu} (left: yellow numbers). Synapses recording a high correlation are kept (right: bold numbers). Synapses with insufficient correlation are changed to a different pre-synaptic neuron (right: yellow numbers). In the lower left an established synaptic connection has been replaced (right: italic yellow) by a random new one.
    \label{fig:topoplast} }}
\end{figure}

Fig.~\ref{fig:topoplast} visualizes the basic concept. It presents an example where each row of synapses gets input from a group of neurons with similar, related information, e.g.\ part of an upstream layer or a subset of neighboring neurons within the layer. The \gls{ppu} assigns random addresses to the synapses of such a row (rows two and three in the example). While the network emulation is continuously operating the synapses in each column measure the correlation between the \gls{nmda} \gls{pp}s and their input. The \gls{ppu} monitors these correlation measurements and tags synapses with high correlation results to be established as working synapses, i.e. assigns them a non-zero weight (bold numbers) while it reassigns new random pre-synaptic input to the synapses showing weak correlation numbers. In addition, it can also reassign previously established synapses if their weight has fallen below a threshold, i.e. if their correlation has weakened over time.

Although a similar net result could be achieved by starting with a fully connected network and pruning unused connections by \gls{stdp}, a much larger number of synapses would be needed initially and a subsequent re-mapping of the remaining non-zero synapses onto the hardware would be necessary to realize any benefit form the pruning.

It is possible to route the post-synaptic firing signal of one compartment, for example the soma, to synapses of different compartments. This allows to implement the functional analogy of back-propagating action potentials from the soma to the dendrites. In a supervised learning scenario this may be used to relate different kinds of teacher signals that modulate somatic firing with the synaptic composition of the neuron.
Similar models to implement biologically plausible back-propagation learning schemes have been proposed\cite{schiess2016}.
Due to the acceleration factor of 1000 the chip can test a multitude of possible synaptic configurations per dendrite while still being faster than biological real time.

\section{outlook\label{sec:outlook}}

This paper presents extensions to the BrainScaleS~2 neuron model for non-linear dendrites and structured neurons using multiple compartments. The purpose of these extensions is not the emulation of the full three-dimensional structure of the neuron\cite{bluebrain05} but its reduction to a minimal electronic model which captures the essential features of such a multi-compartment structure with active, non-linear dendrites. 
We think that the non-linearity created by the different kinds of spikes in combination with a flexible multi-compartment structure will significantly extend the capability of the BrainScaleS~2 system  to help investigate the intersection between biologically inspired hypotheses of information processing\cite{larkum2013}, machine-learning derived approaches\cite{schiess2016} and the efficient implementation of these kinds of processing in dedicated hardware systems.

The ability of a single neuron to act as a coincidence detector and the availability of somatic spike information at distal dendrites is expected to facilitate the mapping of established machine learning approaches to large-scale spiking systems.
Future computing based on neuromorphic hardware might also benefit from the presented level of biological realism, since it could help in creating efficient local learning strategies derived from the biological example. Most likely, if these strategies are identified and proven, the neuromophic systems could be simplified again. Not all of the biological features implemented in the presented model will be needed for each application.

The fully parallel and accelerated nature of the system supports a fast investigation of spiking systems with highly different time scales of neural and plasticity dynamics.
Due to its analog implementation it will keep the advantages of neuromorphic hardware like low power-consumption and robustness against localized defects. 
Since it is spike based and continuous time it might be useful for studying spatio-temporal problems.

The multi-compartment extensions will not change the power-consumption of the \gls{hicann} chip significantly. The energy needed per synaptic transmission depends strongly on network topology and activity and is of the order of magnitude of \SI{10}{\pico\joule}. The area used by the extensions is less than \SI{200}{\micro\meter$^2$} per neuron compartment. This is approximately \SI{0.5}{\percent} of the total area used by neuron and synapse circuits.

The introduced extensions of the BrainScaleS neuron model capture essential features of complex biological structures. Although we do not yet fully understand the purpose and the function of the biological details we are optimistic that the presented models will allow insight into the functional possibilities of multi-layered networks built from multi-compartment neurons possessing non-linear active dendrites. By evaluating the behavior of such networks using a multitude of possible plasticity schemes, we expect to gain insight into which features are relevant for functional performance. Future hardware generations might utilize these insights for systems that incorporate novel nano-electronic components.

The presented circuits have all been implemented in silicon using a \SI{65}{nm} low-power \gls{cmos} technology and are currently being manufactured. As soon as funding allows they will be integrated in the wafer-based BrainScaleS~2 system which will then combine the speed of accelerated neuromorphic hardware with the substantial network size achievable by wafer-scale integration. The presented circuits and concepts should also be transferable to smaller process geometries.

In the meantime, a single chip implementation will soon be available to all interested researchers as an experimental platform for ideas inspired by biology and machine learning and to prepare the ground for future, non-Turing computing substrates.

\section{Acknowledgments}
The authors wish to express their gratitude to Andreas Gr\"ubl, Andreas Hartel, Syed Aamir, Christian Pehle, Korbinian Schreiber, Sebastian Billaudelle, Gerd Kiene, Matthias Hock, Simon Friedmann and Markus Dorn for their participation in the development of the BrainScaleS~2 \glspl{asic} and system.\\
They also want to thank their collaborators Sebastian H\"oppner from TU Dresden and Tugba Demirci from EPFL Lausanne for their contributions to the BrainScaleS~2 prototype \gls{asic}.\\
This work has received funding from the European Union
Seventh Framework Programme ([FP7/2007-2013]) under
grant agreement no 604102 (HBP), 269921 (BrainScaleS),
243914 (Brain-i-Nets), the Horizon 2020 Framework Programme
([H2020/2014-2020]) under grant agreement 720270 (HBP)
as well as from the Manfred St\"ark Foundation.\\
\section{Author Contribution}
J.S. created the concept, designed the circuits and wrote and edited the manuscript except for Section~\ref{sec:results}, L.K. conceived the simulations jointly with P.M, performed the simulations and prepared the results, P.M. wrote Section~\ref{sec:results} and edited the manuscript and K.M. gave conceptual advice.

\end{document}